\documentclass[10pt, conference, compsocconf]{IEEEtran}
\IEEEoverridecommandlockouts
\ifCLASSINFOpdf
\else
\fi

\usepackage{url}
\usepackage{amsfonts}%
\usepackage{amsmath}%
\usepackage{amssymb}%
\usepackage{graphicx}
\usepackage{booktabs}
\usepackage{bm}
\usepackage{balance}

\newcommand{\shortcite}[1]{\cite{#1}}
\newcommand{\commentout}[1]{}
\newcommand{\eat}[1]{}

\newcommand{\bc}{{\bf c}}

\newcommand{\bv}{{\bf v}}

\newcommand{\bx}{{\bf x}}

\newcommand{\by}{{\bf y}}

\newcommand{\bell}{{\bm \ell}}

\newcommand{\realset}{\mathbb{R}}

\newcommand{\abs}[1]{\left|#1\right|}

\newcommand{\sgn}{\mathrm{sgn}}

\newcommand{\transpose}{^\mathsf{\scriptscriptstyle T}}

\begin{document}
%
\title{Conditional Anomaly Detection with Soft Harmonic Functions}




%
\author{\IEEEauthorblockN{Michal Valko\IEEEauthorrefmark{1}$^0$\thanks{$^0$The research work presented in this paper was done when Michal Valko was a
PhD student at the University of Pittsburgh and was supported by funds
from NIH awarded to University of Pittsburgh.},
Branislav Kveton\IEEEauthorrefmark{2},
Hamed Valizadegan\IEEEauthorrefmark{3}, 
Gregory F. Cooper\IEEEauthorrefmark{4} and
Milos Hauskrecht\IEEEauthorrefmark{5}}
\IEEEauthorblockA{\IEEEauthorrefmark{1}INRIA Lille - Nord Europe, SequeL project, 40 avenue Halley, Villeneuve d'Ascq, France, e-mail: michal.valko@inria.fr}
\IEEEauthorblockA{\IEEEauthorrefmark{2}Technicolor Labs, Palo Alto, California, USA, e-mail: branislav.kveton@technicolor.com}
\IEEEauthorblockA{\IEEEauthorrefmark{3}Computer Science Department, University of Pittsburgh, Pennsylvania, USA, e-mail: hamed@cs.pitt.edu}
\IEEEauthorblockA{\IEEEauthorrefmark{4}Department of Biomedical Informatics, University of Pittsburgh, Pennsylvania, USA, e-mail: gfc@pitt.edu}
\IEEEauthorblockA{\IEEEauthorrefmark{5}Computer Science Department, University of Pittsburgh, Pennsylvania, USA, e-mail: milos@pitt.edu}
}


\maketitle

\begin{abstract}
In this paper, we consider the problem of conditional anomaly detection that aims to identify data instances with an unusual response or a class label.  We develop a new non-parametric approach for conditional anomaly detection based on the soft harmonic solution, with which we estimate the confidence of the label to detect anomalous mislabeling. We further regularize the solution to avoid the detection of  isolated examples and  examples on the boundary of the distribution support. We demonstrate the efficacy of the proposed method on several synthetic and UCI ML datasets in detecting unusual labels when compared to several baseline approaches. We also evaluate the performance of our method on a real-world electronic health record dataset where we seek to identify unusual patient-management decisions.

\end{abstract}


%
\IEEEpeerreviewmaketitle

\section{Introduction}
\label{sec:Introduction}

\hyphenation{ana-ly-sis}

Anomaly detection is the task of finding unusual elements in a set of observations. 
Most existing anomaly detection methods in data analysis are unconditional and look for outliers with respect to all data attributes
\cite{markou2003novelty,markou2003noveltya}. Conditional anomaly detection (CAD) \cite{chandola2009anomaly, hauskrecht2007evidence-based, valko2008conditional} is the problem of  detecting unusual values for a subset of variables given the values of the remaining variables. In other words, one set of variables defines the context in which the other set is examined for anomalous values. 

CAD can be extremely useful for detecting unusual behaviors, outcomes, or unusual attribute pairings in many domains \cite{das2008anomaly}.  
Examples of such problems are the detection of unusual actions or outcomes in medicine \cite{hauskrecht2007evidence-based, valko2008conditional, hauskrecht2010conditional}, investments \cite{rubin2005auctioning}, law \cite{aktolga2010detecting}, social networks \cite{heard2010bayesian}, politics \cite{kolar2010estimating} and other fields \cite{das2008anomaly}. In all these domains, the outcome strongly depends on the context (patient conditions, economy and market, case circumstances, etc.), hence the outcome is unusual only if it is compared to the examples with the same context. 

In this work, we study a special case of CAD that tries to identify the unusual values for just one target variable given the values of the remaining variables (attributes). The target variable is assumed to take on a finite set of values which we also refer to as labels, because of its similarity to the classification problems. 
 
In general, the concept of anomaly in data in the existing literature is somewhat ambiguous and several definitions have been proposed in the past \cite{markou2003novelty,markou2003noveltya}.
Typically, an example is considered anomalous when it is not expected from some underlying model.
For the practical purposes of this paper, we define the conditional anomaly detection as follows:

\vskip 0.1cm

\begin{quotation}
\noindent {\bf Problem statement} ($\bigstar$): Given a set of $n$ past examples $(\bx_i,y_i)_{i=1}^{n}$ (with possible label noise), identify and rank instances $i$ in  recent $m$ examples $(\bx_i,y_i)_{i=n+1}^{n+m}$ that are unusual. 
\end{quotation}
In this statement, we do not assume that the labels $\{y_i\}_{i=1}^{n}$ are perfect; they may also be subject to the label noise.

\begin{figure}
	   \begin{center}
		 \includegraphics[width=\columnwidth, clip, viewport=71 271 555 526]{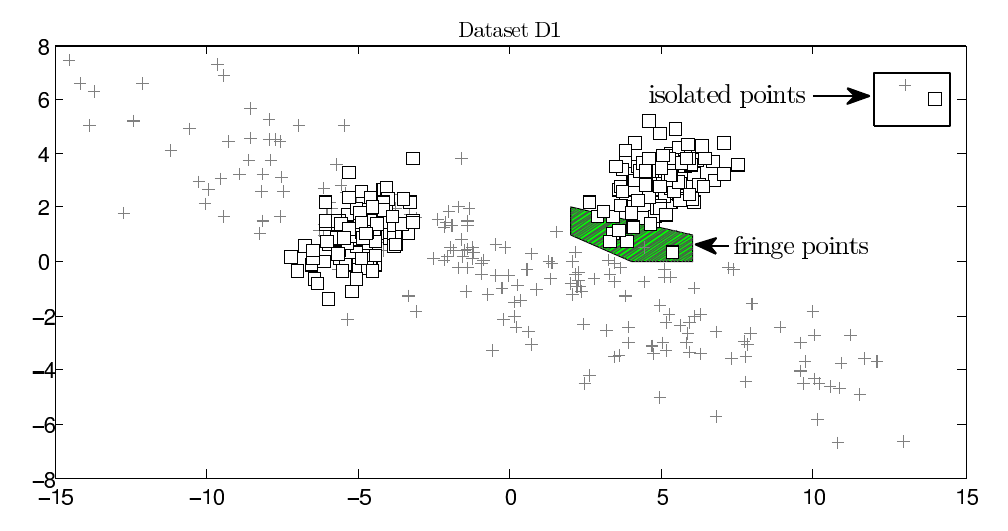}
	   \caption{Challenges for CAD --- the squares and the plus signs represent the examples from the two different classes: 1) {\bf Fringe points} are the examples on the boundary of the class distribution support
		   2) {\bf Isolated points} are the examples far from the majority but not within a different class. 
	   }
		\label{fig:fringe}
		\end{center}
		\end{figure}

In order to assess the anomalousness of an example, we typically output an anomaly score. One way to define the score is to use a probabilistic model $M$ and calculate the anomaly score as the probability of a different label: $P(y\ne y_i|\bx_i,M)$. However, the probabilistic model $M$ is not known in advance and must be estimated from available data. This may lead to two major complications that are illustrated in Figure~\ref{fig:fringe}:
First, an instance may be far from the past observed data points. Because of the lack of the support for alternative responses, it is difficult to assess the anomalousness of these instances.  We refer to these instances as \emph{isolated points}. Second, the examples on the boundary of the class distribution support may look anomalous due to their low likelihood. These boundary examples are also known as \emph{fringe points} \cite{papadimitriou2003cross-outlier}.

One approach to the CAD task is to construct a classification model on the past data $(\bx_i,y_i)_{i=1}^n$ and apply it to $(\bx_i, y_i)_{i=n+1}^{n+m}$ to check if the assigned labels $(y_i)_{i=n+1}^{n+m}$ are correct. However, in that way we would 
disregard the labels $(y_i)_{i=n+1}^{n+m}$ which are available and can be utilized to improve the performance of
the CAD task by leveraging the interaction between the labels of past 
examples and the new observed examples.

Because the underlying conditional distribution of the data is unknown, a non-parametric approach that looks for the label consistency of the instances on their neighborhood (e.g., $k$-nearest neighbor or $k$-NN) can be very useful \cite{papadimitriou2003cross-outlier}. 
One problem with relying on models such as $k$-NN is that they fail to detect clusters of anomalous instances. 


In this paper, we develop and present a new non-parametric method to tackle CAD and its challenges.
Our method relies on the similarity graph of instances and attempts to assess whether the response to an input variable agrees with responses of data points in its neighborhood using propagation of labeling information through the graph. 
Our method differs from typical local neighborhood methods in two important aspects.
First, it respects the structure of the manifold and accounts for more complex interactions in the data. Second, it solves the problem of isolated and fringe points by decreasing the confidence in predicting an opposite label for such points through regularization.

Similar to other graph-based approaches for propagation of information (e.g., semi-supervised learning), the solution to our approach demands the computation of the inverse of the similarity matrix that can be challenging for large number of instances. We address this problem by proposing a method for building a smaller backbone graph that approximates the original graph.

In summary, the main contributions of this paper are:

\begin{itemize}
	\item We introduce a label propagation approach on the data similarity graph for conditional anomaly detection to estimate the confidence of the labels.
\item We propose a specific regularization that avoids unconditional outliers and fringe points  (Figure~\ref{fig:fringe}).
	\item We present a compact computation of unconstrained regularization to account for the approximate backbone graph with different node weights.
	\item  We propose a scaling approach that adjusts for multi-task predictions and makes anomaly scores comparable.
	\item  We verify the efficacy of the proposed algorithm on some synthetic datasets, several UCI datasets and a challenging real-world dataset of patients' health records.
	\end{itemize}

In the following, we first review the related work in Section~\ref{sec:RelatedWork},
and label propagation on the data similarity graph in Section~\ref{sec:UnconstrainedRegularization}. In Section~\ref{sec:CAD}, we adopt the label propagation on the data similarity graph for CAD problem and propose a regularization
that addresses the isolated and fringe points problems.
Next, in Section~\ref{sec:BackboneGraph}, we show how to create a smaller backbone graph to deal
with more than a few thousand examples.
We report the results of our approach on the synthetic datasets, 
UCI ML datasets, and
a real-world medical dataset in Section~\ref{sec:Experiments}.

\section{Background}
\label{sec:Background}

\subsection{Related work}
\label{sec:RelatedWork}

While traditional anomaly detection has been studied for a long time \cite{markou2003novelty,markou2003noveltya}, the methods for CAD are relatively new and the research in this area is only emerging  \cite{chandola2009anomaly,hauskrecht2007evidence-based,song2007conditional}. 
Typically, the existing CAD methods adopt and use classification methods, such as generative models \cite{hauskrecht2007evidence-based,song2007conditional, valko2008distance}, or max-margin classifiers \cite{valko2008conditional,hauskrecht2010conditional}. In Section~\ref{sec:Experiments}, we compare our method to these approaches.

The work on CAD,  when the target variable is restricted to discrete
values only,  is closely related to mislabeling detection \cite{brodley1999identifying} and  cross-outlier detection \cite{papadimitriou2003cross-outlier}. The objective of mislabeling detection is to  1) make a yes/no decision on whether the examples are mislabeled, and 2) improve the classification accuracy by removing the mislabeled examples.

Brodley et~al.~\cite{brodley1999identifying} used different classification approaches to remove mislabeled samples including single and
ensemble classifiers. Bagging and boosting are applied in~\cite{Verbaeten-2003-MisLabeled}.  
Jiang et~al.~\cite{Jiang-2004-Mislabeled} used an ensemble of neural nets to enhance the performance of a $k$-NN classifier. 
Sanchez et~al.~\cite{Sanchez-2003-Mislabeled} introduced several other $k$-NN based approaches including \textit{depuration}, \textit{nearest centroid neighborhood} (NCN) and \textit{iterative} $k$-NCN.
Finally, Valizadegan and Tan~\shortcite{valizadegan2007kernel} introduced an objective function based on the weighted nearest neighbor approach and solved it with Newton method. 

The main difference between mislabeling detection and CAD is that mislabeling detection identifies and removes the mislabeled examples in order to learn better classifiers, while CAD is interested in ranking examples according to the severity of conditional anomalies in data.
This is the main reason our evaluations in Section~\ref{sec:Experiments} measure the rankings of the cases being anomalous and not the improved classification accuracy when we remove them. Nevertheless, we do compare (Sections~\ref{sec:SyntheticData} and~\ref{sec:UCIMLDatasets}) to the methods typically used in mislabeling detection.

Papadimitriou and Faloutsos~\shortcite{papadimitriou2003cross-outlier}
define cross-outliers as examples that seem normal when considering the distribution of examples from the assigned class, but are abnormal when considering the examples from the other class.  For each sample $(\bx,y)$, they compute two statistics based on the similarity of $\bx$ to its neighborhood from the samples belonging to class $y$ and samples not belonging to class $y$. An example is considered anomalous if the first statistic is significantly smaller than the second statistic. However, their method  is not very robust to fringe points (Figure~\ref{fig:fringe}) \cite{papadimitriou2003cross-outlier}. The conditional anomaly approach developed in this paper addresses this problem.

\subsection{Notation}
\label{sec:Notation} 
We use the following notation throughout the paper. Let $(\bx_i,y_i)_{i=1}^{n+m}$ be the collection of $n$ past and $m$ recent observed examples. Without loss of generality, we limit $y$ to binary class labels, i.e.,~$y\in\{\pm1\}$.
Let $G$ be the similarity graph constructed on the nodes $\{\bx_i\}_{i=1}^{n+m}$ with weighted edges $W$.
The entries $w_{ij}$ of $W$ encode the pairwise similarities between $\bx_i$ and $\bx_j$.
We denote by $\mathcal{L}(W)$ the unnormalized graph Laplacian defined as $\mathcal{L}(W)= D - W$
where $D$ is a diagonal matrix whose entries are given by $d_{ii} = \sum_j w_{ij}$.
\eat{where $\bell$ denotes the vector of predictions,}


\subsection{Label Propagation}
\label{sec:UnconstrainedRegularization}
Label propagation on the graph is widely used for semi-supervised learning (SSL).
The general idea is to assume the consistency of labels among the data which are
1) close to each other and 2) lie on some structure (a manifold or a cluster).
The two examples are the \emph{consistency method} of
Zhou~et~al.~\shortcite{zhou2004learning}
and the \emph{harmonic solution} of
Zhu~et~al.~\shortcite{zhu2003semi-supervised}.
The inference of the labels 
by the approach of
Zhu~et~al.~\shortcite{zhu2003semi-supervised}
can be interpreted as a random walk on  $G$ with the transition matrix $P = D^{-1}W$.
The harmonic solution satisfies the \emph{harmonic} property $\ell_i = \frac{1}{d_{ii}}
\sum_{j \sim i} w_{ij} \ell_j$.\footnote{$j\sim i$ denotes that $j$ and $i$ are neighbors in  $G$}

Harmonic solution and consistency method are the instances of a bigger class of the optimization problems called
the unconstrained regularization \cite{cortes2008stability}. In the transductive setting, the unconstrained regularization searches for soft (continuous) label assignment such that it maximizes fit to the labeled data and penalizes for not following the manifold structure:

\begin{equation}
  \bell^\star = \min_{\bell \in \realset^n} \
  (\bell - \by)\transpose C (\bell - \by) + \bell\transpose K \bell,
	\label{eq:unconstrained regularization}
\end{equation}

\noindent
where $K$ is a symmetric regularization matrix and $C$ is a symmetric matrix of empirical weights.
$C$ is  usually diagonal and the diagonal entries often equal to some fixed constant $c_l$ for
the labeled data and $c_u$ for the unlabeled.
In a SSL setting,
$\by$ is a vector of pseudo-targets such that $y_i$ is the label of the $i$-th example when the example is labeled,
and $y_i = 0$ otherwise. Many methods can be derived from \eqref{eq:unconstrained regularization}.
For example, for the (hard) harmonic solution $K = \mathcal{L}(W)$, $c_l = \infty$, and $c_u = 0$.
Consistency method has $K$ equal to the normalized graph Laplacian $K = I-D^{-1/2}WD^{-1/2}$
and $c_u = c_l$ is set to a non-zero constant. The appealing property of \eqref{eq:unconstrained regularization} is that its solution can be computed in closed form as follows~\cite{cortes2008stability}:

\begin{equation}
\bell^\star  = (C^{-1}K+I)^{-1}\by
\label{eq:closed form}
\end{equation}

\eat{\cite{valko2010online} $K = L + \gamma_g I$
For regularized, is the regularized Laplacian of the
similarity graph.
}

\section{Methodology}
\label{sec:Methods}

In this section we show how to solve the CAD problem $(\bigstar$) using label propagation on a data similarity graph
and how to compute the anomaly score.
In particular, we will build on the harmonic solution approach (Section~\ref{sec:UnconstrainedRegularization})
and adopt it for CAD in the following ways:
	1) show how to compute the confidence of mislabeling, %
	2) add a regularizer to address the problem of isolated and fringe points, %
	3) use soft constraints to account for a fully labeled setting, and %
	4) describe a compact computation of the solution from a quantized backbone graph. %

\subsection{Conditional Anomaly Detection}
\label{sec:CAD}
The label propagation method described in Section~\ref{sec:UnconstrainedRegularization} can be applied to CAD by considering all observed data as labeled examples with no unlabeled examples. The setting for matrix $C$ is dependent on the quality of the past observed data. If the labels of the past observed data (or any example from the recent sample) are guaranteed to be correct, we set the corresponding diagonal elements of $C$ to a large value to make their labels fixed. Notice that specific domain techniques can be used to make sure that the collected examples from the past observed data have correct labels. In this paper, we assume that we do not have access to such prior 
knowledge and therefore, the observed data are also subject to label noise.

We now propose a way to compute the anomaly score from~\eqref{eq:closed form}.
The output $\bell^\star$ of \eqref{eq:unconstrained regularization} for the example $i$ can be rewritten as:
\begin{equation}
\ell_i^\star = |\ell_i^\star| \times \mathrm{sgn}(\ell_i^\star)
\label{eq:absform}
\end{equation}
\noindent SSL methods use $\mathrm{sgn}(\ell^\star_i)$ in  \eqref{eq:absform}
as the predicted label for $i$. For an unlabeled example, when the value of $\ell_i$ is close to $\pm 1$,
then the labeling information that was propagated to it is more consistent.
Typically, that means that the example is close to the labeled examples of the respective class.
The key observation, which we exploit in this paper, is
that we can interpret $|\ell^\star_i|$ as a confidence in the label.
Our situation differs from SSL, as all our examples are labeled and we aim to assess the confidence of \emph{already labeled} example.
Therefore, we define the \emph{anomaly score} as the
absolute difference between the actual label $y_i$ and the inferred soft label $\ell_i$:

\begin{equation}
s_i = |\ell_i^\star - y_i|.
\label{eq:anomalyscore}
\end{equation}

We will now address the problems illustrated in Figure~\ref{fig:fringe}.
Recall that the isolated points are the examples
that are (with respect to some metric)  far from the majority of the data.
Consequently, they are surrounded by few or no nearby points.
Therefore, no matter what their label is, we do not want to report them as conditional anomalies. In other words,
we want CAD methods to assign them a low anomaly score.
Even when the isolated points are far from the majority data, they  still can be orders
of magnitudes closer to the data points with the opposite label.
This can make a label propagation approach falsely confident about that example being a conditional anomaly.
In the same way we do not want to assign a high anomaly score  to fringe points just because they lie on a distribution boundary.
To tackle these problems we set $K = \mathcal{L}(W) + \gamma_gI$, where we diagonally regularize the graph Laplacian.
Intuitively, such a regularization lowers the confidence value $|\bell^\star|$ of all examples; however it reduces the confidence score of far outlier points relatively more. To see this, notice (Section~\ref{sec:Dataset}) that
the similarity weight metric is an exponentially decreasing function of the Euclidean distance.
In other words, such a regularization can be interpreted as a label propagation on the graph with an extra
sink. The sink is an extra node in $G$ with label $0$ and every other node connected to it
with the same small weight $\gamma_g$. The edge weight of $\gamma_g$ affects the isolated points more than other points because their connections to other nodes are small.


In the fully labeled setting, the \textit{hard} harmonic solution degenerates to the weighted $k$-NN.
In particular, the hard constraints of the harmonic solution do not allow the labels spread beyond other labeled examples.
However, despite the fully labeled case, we still want to take the advantage of the manifold structure.
To alleviate this problem we allow labels to spread on the graph by
using soft constraints in the unconstrained regularization problem \eqref{eq:unconstrained regularization}.
In particular, instead of $c_l=\infty$ we set $c_l$ to a finite constant and we also set $C = c_l I$.
With such a setting of $K$ and $C$, we can solve \eqref{eq:unconstrained regularization}
using \eqref{eq:closed form} to get:

\begin{align}
\bell^\star & = \left(\left(c_l I\right)^{-1}\left(\mathcal{L}(W)+\gamma_g\right) + I\right)^{-1}\by \\
      & = \left(c_l^{-1}\mathcal{L}(W)+\left(1+\frac{\gamma_g}{c_l}\right)I\right)^{-1}\by.
			\label{eq:had}
\end{align}

\noindent To avoid computation of the inverse,\footnote{due to numerical instability} we calculate \eqref{eq:had} using the following system of linear equations:

\begin{equation}
\left(c_l^{-1}\mathcal{L}(W)+\left(1+\frac{\gamma_g}{c_l}\right)I\right) \bell^\star = \by
\label{eq:had:sle}
\end{equation}

\eat{
[Ie. $\left(c_l^{-1}\mathcal{L}(W)+\left(1+\frac{\gamma_g}{c_l}\right)I\right) \backslash \by$ with \textsc{Matlab}]
}
\noindent
We then plug the output of \eqref{eq:had:sle} into \eqref{eq:anomalyscore} to get the anomaly score.
We will refer to this score as the SoftHAD score.
Intuitively, when the confidence is high but $\mathrm{sign}(\ell_i^\star) \ne y_i$, we will consider the label $y_i$ of the case
$(\bx_i, y_i)$ conditionally anomalous.
\eat{

We can perform the method described in Section \ref{sec:Methods} using traditional batch setting.
\emph{This does not scale and does not adapt.}

Alternatively, we can let trivially modify it to the online version and
let the graph grow with every new example. Specifically, every step $t$ we add $\bx_t$
to the graph and recompute (\ref{eq:had:sle}). \emph{This does adapt but does not scale.}

The regularized HS  is
an offline learning algorithm. This algorithm can be made na\"{i}vely online,
by taking each new example, connecting it to its neighbors,
and recomputing the HS. Unfortunately, this na\"ive implementation has computational complexity
$O(t^3)$ at step $t$, and computation becomes infeasible
as more examples are added to the graph.
}

\subsection{Backbone Graph}
\label{sec:BackboneGraph}

The computation of the system of linear equations \eqref{eq:had:sle}
scales with complexity\footnote{The complexity can be further improved to $O(n_u^{2.376})$ with
the Coppersmith-Winograd algorithm.} $O(n^3)$.
This is not feasible for a graph with more than several thousand nodes.
To address the problem, we use \emph{data quantization} \cite{gray1998quantization}
and sample a set of nodes from the training data to create $G$.
We then substitute the nodes in the graph with a smaller set of $k \ll n$ distinct centroids
which results in $O(k^3)$ computation.

We improve the approximation of the original graph with the backbone graph,
by assigning different weights to the centroids.
We do it by computing the multiplicities (i.e., how many nodes each centroid represents).
In the following we will describe how to modify
 \eqref{eq:had:sle} to allow for the computation with multiplicities.

Let $V$ be the diagonal matrix of multiplicities with $v_{ii}$ being
the number of nodes that centroid $\bx_i$ represents.
We will set the multiplicities according to the empirical prior.
Let $W^V$ be the compact representation of the matrix $W$ on $G$,
where each node $\bx_i$ is replicated $v_{ii}$ times.
Let $L^V$ and $K^V$ be the graph Laplacian and regularized
graph Laplacian of $W^V$. Finally, let $C^V$ be the $C$ in \eqref{eq:unconstrained regularization} with the adjustment for the multiplicities. $C^V$ accounts for the fact that we care about ``fitting'' to train data according to the train data multiplicities.
Then:

\begin{align*}
W^V &= VWV\\
L^V &= \mathcal{L}(W^V)\\
K^V &= L^V + \gamma_gV\\
C^V &= V^{1/2}CV^{1/2}	
\end{align*}

\noindent
\noindent The unconstrained regularization \eqref{eq:unconstrained regularization}
now becomes:

\begin{equation}
  \bell^{V\star} = \min_{\bell \in \realset^n} \
  (\bell - \by)\transpose C^V (\bell - \by) + \bell\transpose K^V\bell
	\label{eq:unconstrained regularization:mul}
\end{equation}

\noindent and subsequently \eqref{eq:had} becomes:

\begin{eqnarray*}
 \bell^{V\star}  & = & \left(\left(C^V\right)^{-1}K^V+I\right)^{-1}\by \\
      & = & \left( V^{-1/2}C^{-1}V^{-1/2} (L^V + \gamma_gV)  +I\right)^{-1}\by \\
\label{eq:closed form:mul}
      & = & \left(\left(c_lV\right)^{-1}  (L^V + \gamma_gV)  +I\right)^{-1}\by \\
      & = & \left(1/c_l V^{-1}  L^V + c_l\gamma_g  +I\right)^{-1}\by
\end{eqnarray*}

\noindent With these adjustments the anomaly score that accounts
for the multiplicities is equal to $|\bell^{V\star} - \by|$.

\begin{figure*}
\begin{center}
\includegraphics[width=2\columnwidth, viewport=76 26 708 402]{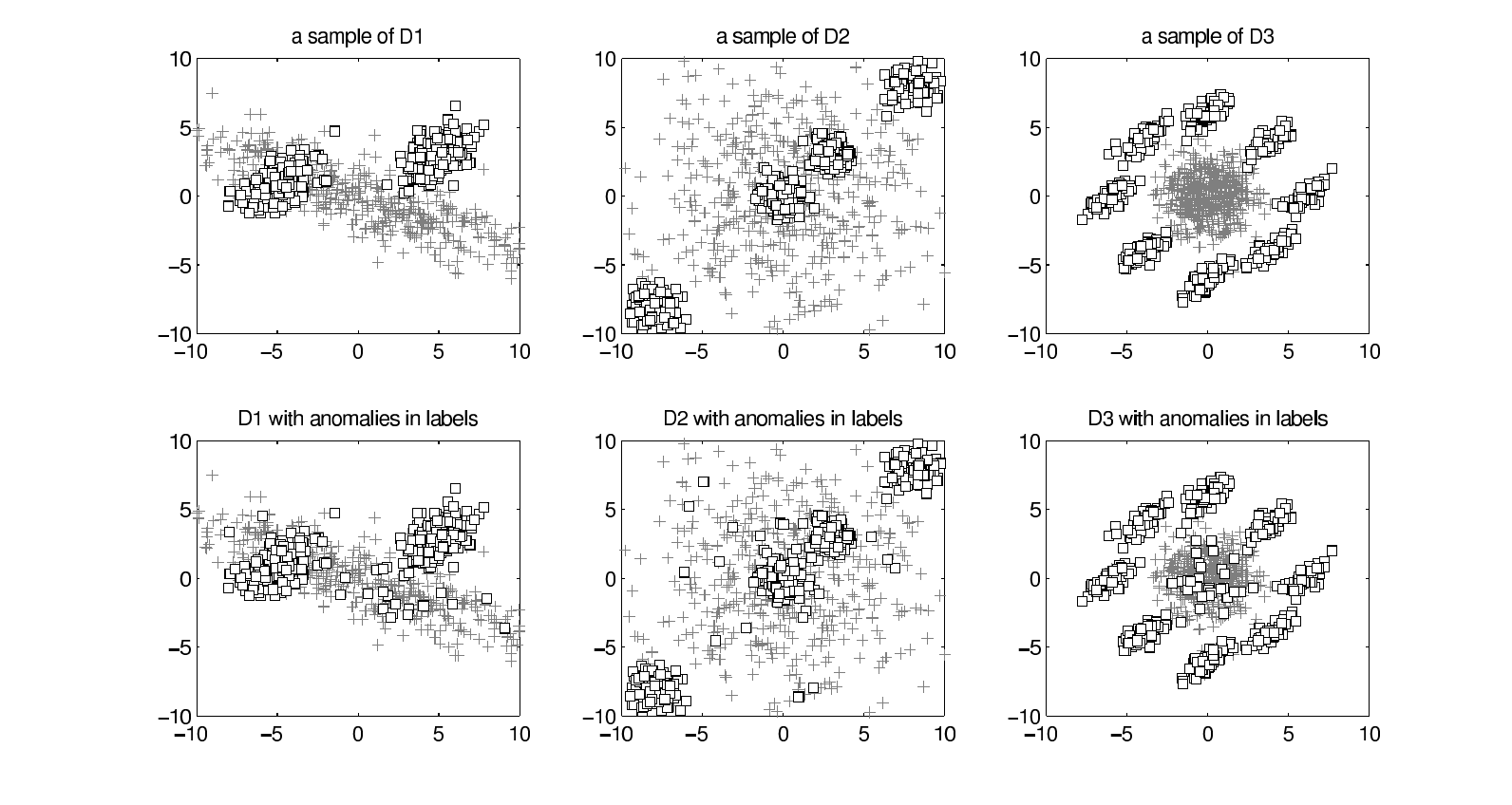}
\caption{Synthetic Data. Top: A sample of datasets D1, D2, and  D3. Bottom: Synthetic datasets after changing the labels of 3\% of the examples.}
\label{fig:syn_ano}%
\end{center}
\end{figure*}

\section{Experiments}
\label{sec:Experiments}

To evaluate the performance of our SoftHAD method, we compare  it to the following baselines:
\begin{itemize}
\item 1-class SVM approach in which we cover each class by a separate 1-class SVM \cite{scholkopf1999estimating}
with RBF kernel and the anomaly score equals to the distance of the example from the learned boundary of its own class. 
This method is an example of the traditional anomaly detection method adopted for CAD.
\item Quadratic discriminant analysis (QDA) model \cite{hastie2001elements}, where we model each class by a multivariate Gaussian,
and the anomaly score is the class posterior of the opposite class.
\item SVM classification model \cite{vapnik1995nature} with RBF kernel where we consider an example anomalous if it falls far on
the opposite side of the decision boundary. This method is an example of the classification method adopted for CAD and
was used  by 
Valko~et~al.~\shortcite{valko2008conditional}.
\item Weighted $k$-NN approach \cite{hastie2001elements} that uses the same weight metric $W$ as SoftHAD, but relies 
only on the labels in the local neighborhood and does not account for the manifold structure.
\end{itemize}

\subsection{Synthetic data}
\label{sec:SyntheticData}

The evaluation of a CAD is a very challenging
task when the true model is not known. Therefore, we first
evaluate and compare the results of different CAD methods
on three synthetic datasets (D1, D2, and D3) with known underlying models
that let us compute the true anomaly scores.

We show the three datasets we used in our experiments in Figure~\ref{fig:syn_ano}.
All datasets are modeled as the mixtures of multivariate Gaussians
and the class densities we used to generate 
these datasets vary in locations, shapes and mutual overlaps. 
Dataset D1 is similar to XOR type of data with one of the classes modeled by a single elongated Gaussian.
In D2, the classes overlap and the form of D3 is close to the concentric circles but the clusters are non-overlapping.
For each dataset, we sampled 500 examples from the class $+1$ and 500 examples from the class $-1$ for
the training set and the same number of the examples for the testing set. 
For each experiment we sample the datasets 100 times. 
After the sampling, we randomly switch the class labels for three percent of examples
for both training and testing set. 

\commentout{
We used a dataset shown from Figure~\ref{fig:fringe} to evaluate different
CAD methods. The positive class consists of 200 points sampled from:
\begin{equation*}
\mathcal{N}\left(\left(\begin{array}{r}
0 \\
0 \end{array}
\right)
\left( \begin{array}{rr}
50 & -20 \\
-20 & 10 \end{array}
\right)
\right)
\end{equation*}
\noindent and a point at $[17,8]$. The negative class consists
of 2$\times$100 points each sample from the two normal distributions:
\begin{equation*}
\mathcal{N}\left(\left(\begin{array}{r}
-5 \\
-1 \end{array}
\right)
\left( \begin{array}{rr}
1 & 0.5 \\
0.5 & 1 \end{array}
\right)
\right),  
\mathcal{N}\left(\left(\begin{array}{r}
5 \\
3 \end{array}
\right)
\left( \begin{array}{rr}
1 & 0.5 \\
0.5 & 1 \end{array}
\right)
\right)
\end{equation*}
\noindent and a point at $[15,8]$.
} 



We then calculate the true anomaly score as $$P(y \neq y_i|\bx_i) = 1 - P(y=y_i|\bx_i),$$ reflecting how anomalous the label of the example is with respect to the true model.
Each of the methods outputs a score which orders the examples according to the belief of the anomalous labeling. For each of the CAD methods, we assess how much this ordering is consistent with the
ordering of the true anomaly score. We did it by counting the number of swapped pairs between the true and predicted ordering which is equivalent to the Wilcoxon score or AUC score.\footnote{AUC commonly used for classification is a special case with the true score being $\pm 1$.} The less number of swapped pairs, the higher the agreement score.


Table \ref{tab:syn} compares the agreement scores (or equivalently, the AUCs) of the experiment for all methods. The results demonstrate that our method outperforms all other baselines and comes closest to the order induced by the true model.
We also evaluated the linear versions of SVM and 1-class SVM, but the results were inferior to the ones with the RBF kernel.
\begin{table}[htbp]
  \centering
      \begin{tabular}{rc|c|c}
    \addlinespace
    \toprule
        &  \phantom{a} Dataset \textbf{D1} \phantom{a}  & \phantom{a} Dataset \textbf{D2} \phantom{a}   &  \phantom{a} Dataset \textbf{ D3} \phantom{a}  \\
    \midrule		    
    \emph{QDA} & 73.4\% (2.4) & 48.0\% (1.4) & 54.5\% (1.2) \\
    \emph{SVM} & 59.8\% (4.8) & 58.8\% (5.7) & 50.8\% (2.2) \\
    \emph{1-class SVM} & 50.7\% (1.3) & 52.3\% (1.2) & 59.4\% (1.6) \\
    \emph{w$k$--NN} & 66.6\% (1.4) & 64.6\% (1.3) & 60.5\% (1.5) \\
		\emph{SoftHAD} & 81.3\% (1.8) & 82.4\% (1.6) & 63.0\% (2.9) \\					
	
    \bottomrule
    \end{tabular}%
		\vskip 0.25cm
		\caption{Mean anomaly agreement score and variance (over 100 runs) for CAD methods on the 3 synthetic datasets.}
  \label{tab:syn}%
\end{table}%

\begin{figure*}
\begin{center}
\includegraphics[width=1.7\columnwidth, viewport=0 328 594 627]{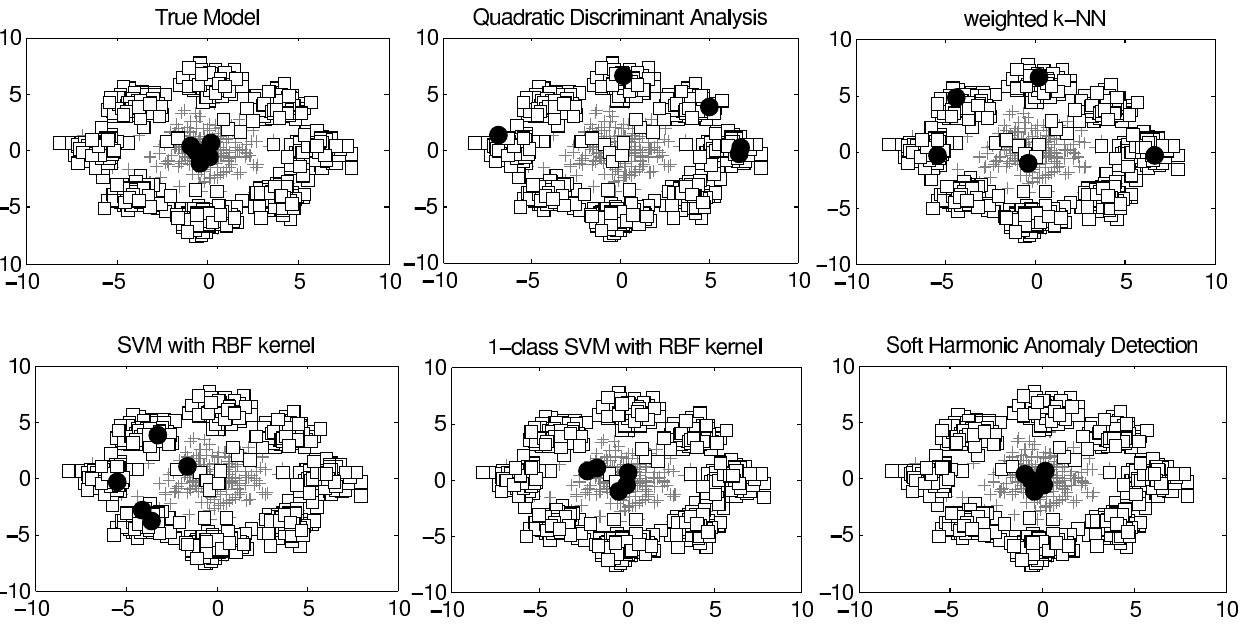}
\caption{Black dots depict the top five conditional anomalies based on the score for each of the methods on D3. The top five conditional anomalies according to the true model are in the middle (top left).}
\label{fig:top_syn_ano}
\end{center}
\end{figure*}

Figure \ref{fig:top_syn_ano} shows the top 5 anomalies identified by each method on D3.
We see that only the soft harmonic method was able to identify the top conditional anomalies, which correspond to the examples with
switched labels in the middle region that carries a lot of counter-support and hence leads to the highest anomaly score.       
\subsection{UCI ML Datasets}
\label{sec:UCIMLDatasets}

We also evaluated our method on the three UCI ML datasets~\cite{asuncion2007uci}, for 
which an ordinal response variable was available to calculate the true anomaly score. In particular, we
selected 1) \emph{Wine Quality} dataset with the response variable \emph{quality}
2) \emph{Housing} dataset with the response variable \emph{median value of owner-occupied homes}
and 3) \emph{Auto MPG} dataset with the response variable \emph{miles per gallon}.
In each of the dataset we scaled the response variable $y_r$ to the $[-1,+1]$ interval
and set the class label as $y := y_r \geq 0$. 
As with the synthetic datasets, we randomly switched the class labels for 
three percent of examples.
The true anomaly score was computed as the absolute difference 
between the original response variable $y_r$ and the (possibly switched) label.
Table~\ref{tab:uci} compares the agreement scores to the true score for all methods on 
(2/3, 1/3) train-test split. Again, we see that SoftHAD either performed the best
or was close to the best method.


\begin{table}[htbp]
  \centering
 \begin{tabular}{rc|c|c}
    \addlinespace
    \toprule
        & \phantom{a}\textbf{Wine Quality}\phantom{a}&\phantom{a} \textbf{Housing} \phantom{a}& \phantom{a}\textbf{Auto MPG}\phantom{a} \\
    \midrule
	   \emph{QDA}& 75.1\% (1.3) & 56.7\% (1.5) & 65.9\% (2.9) \\
   \emph{SVM}  & 75.0\% (9.3) & 58.5\% (4.4) & 37.1\% (8.6) \\
    \emph{1-class SVM}  & 44.2\% (1.9) & 27.2\% (0.5) & 50.1\% (3.5) \\
    \emph{w$k$--NN}         & 67.6\% (1.4) & 44.4\% (2.0) & 61.4\% (2.3) \\
		\emph{SoftHAD}      & 74.5\% (1.5) & 71.3\% (3.2) & 72.6\% (1.7) \\	
    \bottomrule
    \end{tabular}%
		\vskip 0.25cm
		\caption{Mean anomaly agreement score and variance (over 100 runs) for CAD methods on the 3 UCI ML datasets.}
  \label{tab:uci}%
\end{table}%

\subsection{Medical data}
\label{sec:Dataset}

In this experiment, we evaluated our method on the problem of detecting unusual patient-management actions~\cite{hauskrecht2010conditional}.
We asked a panel of clinical experts to judge the outputs of the CAD methods 
for the clinical relevance.  

\subsubsection{Data} We used the data extracted from
electronic health records (EHRs) of 4,486 patients as described in~\cite{hauskrecht2010conditional}. The patients were divided into a train set
(2646 patients) and a test set (1840 patients). Patient records were segmented 
in time (every day of a patient's visit at 8:00am) to obtain 51,492 patient-state instances, such that
30,828 were train and 20,664 test instances. The data in EHRs for these instances 
were then converted into 9,282 features -- a vector representation of the patient state. 
For every patient-state instance we had 749 decision labels (or tasks)  
which were possible lab-order and medication
decisions with true/false values, reflecting whether a particular lab was ordered or a particular medication
was given within a 24-hour period. 

\subsubsection{Evaluation}
We evaluated our CAD method on 222 patient-instance/action pairs. 
We selected these 222 cases such that they represented a wide range of low, medium and high anomaly scores 
according to the baseline SVM method \cite{valko2008conditional,hauskrecht2010conditional}. Each instance/action pair was evaluated by three different clinical experts
determining whether the action is anomalous and whether this anomaly is clinically relevant. 
To assess the example, we used the majority
rule (two out of three experts). 
We then evaluated the quality of CAD methods using the area under the ROC (AUC) metric. 
We compared SoftHAD method to the three baselines
1) weighted $k$--NN on the same graph 2) SVM with RBF kernel 3) 1-class SVM with RBF kernel
described at the beginning of this section.
    
\subsubsection{Parameters for the graph-based algorithms}
\label{sec:Parameters}

To construct $G$, we computed the similarity weights as:
$$w_{ij} = \exp\left[- \left(||\bx_i - \bx_j||_{2,\psi}^2 \right) / \sigma^2 \right],$$

\noindent where $\psi$ is a weighing of the features and $\sigma$ is a length scale parameter.
The reason for the different feature weights is the high dimensionality of the data.
Without any feature scaling, a distance based on 9K features would make any two points almost equidistant
and thus meaningless. Therefore, we weighted the features based on their discriminative power
according to the univariate Wilcoxon score \cite{hanley1982meaning}.
Next, $\sigma$ is chosen so that the graph is reasonably sparse \cite{luxburg2007tutorial}.
We followed \cite{valizadegan2007kernel} and chose $\sigma$ as 10\% of the empirical variance of the Euclidean distances.
Based on the experiments, our algorithm is not sensitive to the small perturbations of $\sigma$; what is important is that the graph does not become disconnected by having all edges of several nodes with weights close to zero.
For each label, we sampled an equal number of positive and negative instances to construct a
$k$-NN graph. We set $k=75$, $c_l = 1$ and varied $\gamma_g$ and the graph size.




\subsubsection{Scaling for multi-task anomaly detection}
\label{sec:Scaling}
\begin{figure}
\includegraphics[width=\columnwidth,clip, viewport=45 308 589 463]{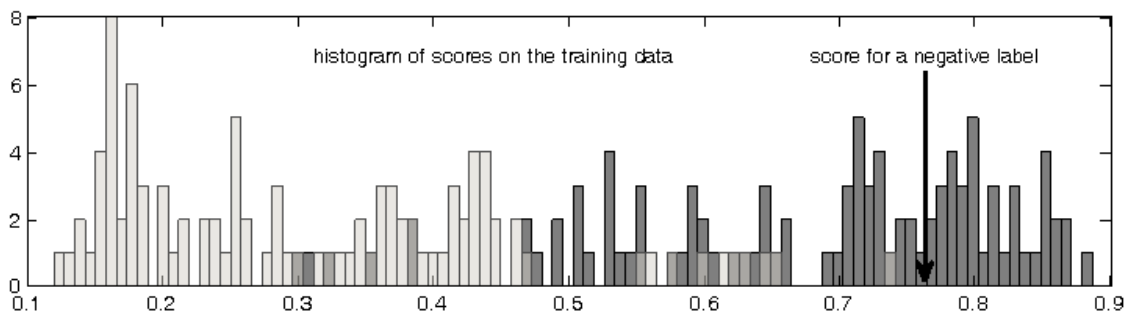}
\includegraphics[width=\columnwidth,clip, viewport=58 316 558 471]{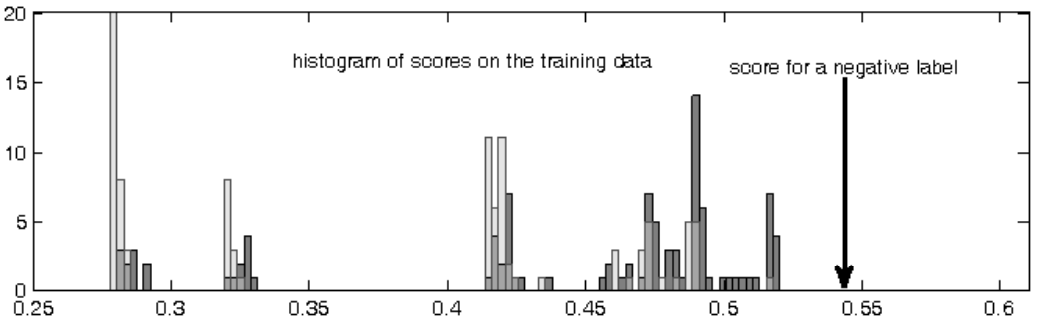}
\caption{Histogram of anomaly scores for 2 different tasks.
The scores for the top and bottom task range from 0.1 to 0.9
and from 0.25 and 0.61, respectively. The arrow in both cases
points to the scores of the evaluated examples, both with negative labels. Despite the score is lower for the bottom
task, we may believe that it is more anomalous because it is
more extreme within the scores for the same task.}
\label{fig:scaling2}%
\end{figure}

So far, we have described CAD only for a single task (anomaly in a single label). 
In this dataset, we have 749 binary tasks (or labels) that correspond
to 749 different possible orders of lab tests or medications.
In our experiments, we compute the CAD score for each task independently.
Figure \ref{fig:scaling2} shows the CAD scores for two of them.
CAD scores close to 1 indicate that the order should be done, while
the scores close to 0 indicate the opposite. 
The ranges for the anomaly scores can vary
among the different labels/tasks, as one can notice in Figure \ref{fig:scaling2}.
However, we want to output an anomaly score which is comparable among the different tasks/labels
so we can set a unified threshold when the system is deployed in practice.
To achieve this score comparability, we 
propose a simple approach, where we take the minimum and the maximum score obtained for the training set
and scale all scores for the same task linearly so that the score after the scaling ranges from 0 to 1.

\subsubsection{Results}
\label{sec:Results}

\begin{figure}
\begin{center}
\includegraphics[width=\columnwidth,clip, viewport=52 265 539 522]{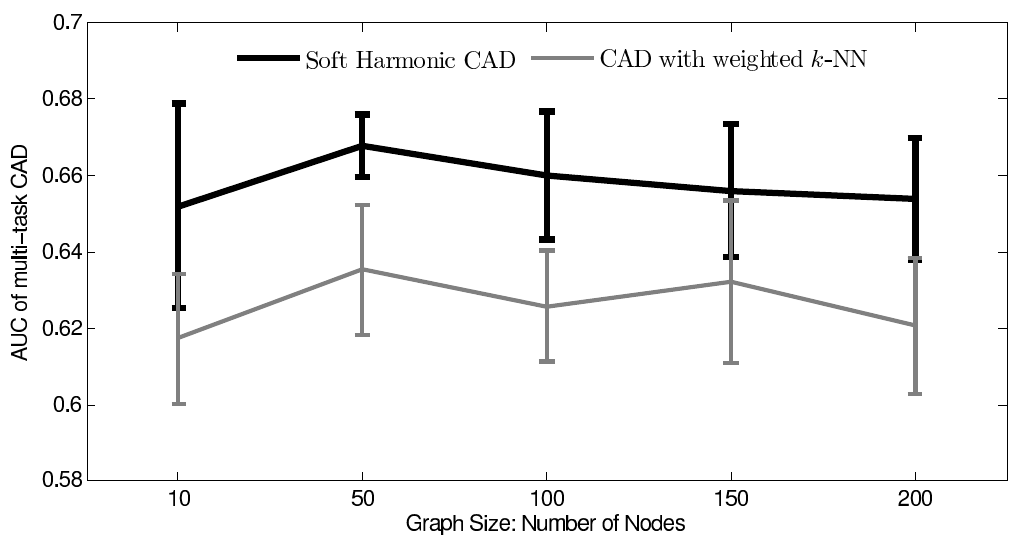}
\caption{Medical Dataset: Varying graph size. Comparison of 1) SoftHAD and 2) weighted $k$-NN on the same graph.}%
\label{fig:samplesize}%
\end{center}
\end{figure}

\begin{figure}
\begin{center}
\includegraphics[width=\columnwidth, clip, viewport=39 251 565 530]{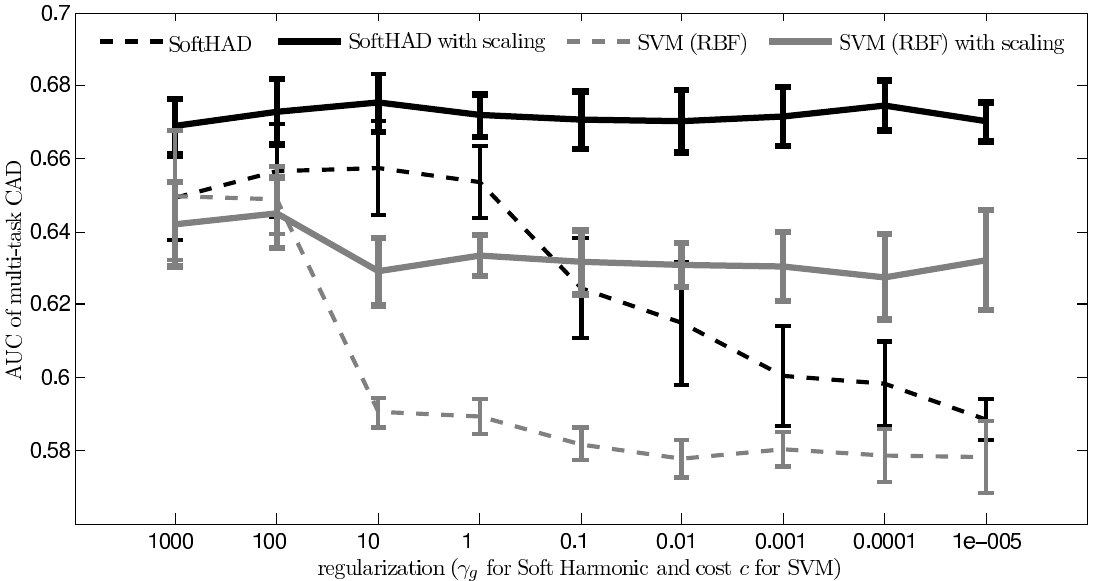}
\caption{Medical Dataset: Varying regularizer 1) $\gamma_g$ for SoftHAD 2) cost $c$ for SVM with RBF kernel.}
\label{fig:gg}%
\end{center}
\end{figure}

In Figure~\ref{fig:samplesize}, we fixed $\gamma_g = 1$ and vary the number of examples we sample from the training set to construct the similarity graph, and also compare it to the weighted $k$--NN. The error bars show the variances over 10 runs.
Notice that the both of the methods are not too sensitive to the graph size.
This is due to the multiplicity adjustment  for the backbone graph (Section~\ref{sec:BackboneGraph}).
Since we use the same graph both for SoftHAD and weighted $k$--NN, we anticipate that we are able to outperform
weighted $k$--NN due to the label propagation over the data manifold and not only within the immediate neighborhood.
In Figure~\ref{fig:gg}, we compare SoftHAD to the CAD using SVM with an RBF kernel for different regularization settings.
We sample 200 examples to construct a graph (or train an SVM) and vary the $\gamma_g$ regularizer (or cost $c$ for SVM).
We outperform the SVM approach over the range of regularizers.
The AUC for the 1-class SVM with an RBF was consistently below 55\%, so we do not show it in the figure.
We also compared the two methods with scaling adjustment for this multi-task problem (Figure~\ref{fig:gg}).
The scaling of anomaly scores improved the performance of both methods and makes the methods less sensitive to the regularization settings.

\section{Conclusion}
\label{sec:Conclusion}
We presented a non-parametric graph-based algorithm for conditional anomaly detection.
Our algorithm goes beyond exploring just the local neighborhood (nearest neighbor approach) 
and uses label propagation on the data manifold structure to estimate the confidence of labeling.
We evaluated our method on synthetic data, where the true model was known to 
confirm that the anomaly score from our method outperforms the others in ordering examples 
according to the true anomaly score. We also presented the evaluation of our method on the real-world data of patient 
health records, where the true model is not know, but when we used the experts in clinical
care to evaluate the severity of our alerts.

In future, we plan to work on the structure anomalies where instead of 
computing the anomaly score independently for each label, we compute it \emph{jointly}.
With such a structured approach we can avoid the necessity of the score scaling.

\section{Acknowledgments} 

This research work was supported by the grants  R21LM009102, R01LM010019, and R01GM088224 
from the NIH and the grant IIS-0911032 from the NSF. Its content is solely the responsibility of the authors and does not necessarily represent the official views of the NIH or NSF.
Moreover, it was also supported by Ministry of Higher
Education and Research, Nord-Pas de Calais Regional Council and FEDER
through the ``contrat de projets {\'e}tat region 2007--2013", and by
PASCAL2 European Network of Excellence. The research leading to these
results has also received funding from the European Community's
Seventh Framework Programme (FP7/2007-2013) under grant agreement n$^{\rm o}$
270327 (project CompLACS).

\balance

\bibliographystyle{IEEEtran}
\bibliography{IEEEabrv,miki}

\end{document}